\documentclass[letterpaper]{article} 
\usepackage{aaai25}  
\usepackage{times}  
\usepackage{helvet}  
\usepackage{courier}  
\usepackage[hyphens]{url}  
\usepackage{graphicx} 
\usepackage{natbib}  
\usepackage{caption} 
\usepackage{algorithm}

\usepackage{newfloat}
\usepackage{listings}
\DeclareCaptionStyle{ruled}{labelfont=normalfont,labelsep=colon,strut=off} 
\floatstyle{ruled}
\newfloat{listing}{tb}{lst}{}
\floatname{listing}{Listing}
\title{MEDSAGE: Enhancing Robustness of Medical Dialogue Summarization to ASR Errors with LLM-generated Synthetic Dialogues}
\author{
    Kuluhan Binici\textsuperscript{\rm 1,\rm 2,\rm 3$\dagger$ \blfootnote{Work done \textsuperscript{$\dagger$}as part of an internship at AICS, 
\textsuperscript{$\mathsection$}while the authors were at AICS.}},
    Abhinav Ramesh Kashyap\textsuperscript{\rm 4$\mathsection$},
    Viktor Schlegel\textsuperscript{\rm 5,\rm 6$\mathsection$},
    Andy T. Liu\textsuperscript{\rm 1$\mathsection$}, 
    \\
    Vijay Prakash Dwivedi\textsuperscript{\rm 7$\mathsection$}, 
    Thanh-Tung Nguyen\textsuperscript{\rm 1$\mathsection$},
    Xiaoxue Gao\textsuperscript{\rm 8}, 
    Nancy F. Chen\textsuperscript{\rm 8},
    Stefan Winkler\textsuperscript{\rm 1,\rm 3}
}
\affiliations{
    \textsuperscript{\rm 1}ASUS Intelligent Cloud Services (AICS)\\ 
    \textsuperscript{\rm 2}SAP\\
    \textsuperscript{\rm 3}National University of Singapore\\
    \textsuperscript{\rm 4}Crayon Software \\
    \textsuperscript{\rm 5}Imperial College London, Imperial Global Singapore\\
    \textsuperscript{\rm 6}University of Manchester, Department of Computer Science\\
    \textsuperscript{\rm 7}Stanford University\\
    \textsuperscript{\rm 8}Institute for Infocomm Research, A*STAR\\
}

\usepackage{bibentry}

\usepackage{algorithmicx}
\usepackage{algpseudocode}
\usepackage{colortbl}
\usepackage{xcolor}
\usepackage{amsmath}
\usepackage{booktabs}
\usepackage{tcolorbox}
\usepackage{subcaption}
\usepackage{soul}

\newcommand{\frameworkname}{\texttt{MEDSAGE}}
\newcommand{\redhl}[1]{\sethlcolor{red!30}\hl{#1}\sethlcolor{white}}
\newcommand{\greenhl}[1]{\sethlcolor{green!30}\hl{#1}\sethlcolor{white}}
\newcommand{\yellowhl}[1]{\sethlcolor{yellow!30}\hl{#1}\sethlcolor{white}}
\newcommand{\rulesep}{\unskip\ \vrule\ }
\newcommand\blfootnote[1]{%
  \begingroup
  \renewcommand\thefootnote{}\footnote{#1}%
  \addtocounter{footnote}{-1}%
  \endgroup
}

\begin{document}

\maketitle

\begin{abstract}
Automatic Speech Recognition (ASR) systems are used to transcribe speech into text, yet the errors they introduce can significantly degrade the performance of downstream tasks like summarization. This issue is particularly pronounced in clinical dialogue summarization, a low-resource domain where supervised data for fine-tuning is scarce, necessitating the use of ASR models as black-box solutions. Employing conventional data augmentation for enhancing the noise robustness of summarization models is not feasible either due to the unavailability of sufficient medical dialogue audio recordings and corresponding ASR transcripts. To address this challenge, we propose MEDSAGE, an approach for generating synthetic samples for data augmentation using Large Language Models (LLMs). Specifically, we leverage the in-context learning capabilities of LLMs and instruct them to generate ASR-like errors based on a few available medical dialogue examples with audio recordings. Experimental results show that LLMs can effectively model ASR noise, and incorporating this noisy data into the training process significantly improves the robustness and accuracy of medical dialogue summarization systems. This approach addresses the challenges of noisy ASR outputs in critical applications, offering a robust solution to enhance the reliability of clinical dialogue summarization.
\end{abstract}

%

\section{Introduction}
Automatic Speech Recognition (ASR) \cite{yu2016automatic} is the task of transcribing speech signals into text, enabling a wide range of applications from voice-activated assistants to automated customer service systems. ASR systems significantly aid various downstream tasks such as dialogue summarization \cite{liu2019topic,zhong2022dialoglm}, where the goal is to distill key information from spoken interactions. However, errors introduced by ASR systems can degrade the performance of these summarization tasks \cite{li2014overview,guo2024ucorrect}, which limits their application in high-stake domains where correctness of the summaries is important. The synthesis of Electronic Medical Records (EMRs) from doctor-patient dialogues~\cite{krishna-etal-2021-generating} is among such summarization tasks where accuracy is critical. Here, ASR errors can lead to inaccuracies in transcribing clinical terms, medication, or procedure names, resulting in erroneous medical notes~\cite{hodgson2016risks}, which can lead to misdiagnosis, incorrect treatment plans, and potentially harmful patient outcomes.

One way to improve dialogue summarization is to improve the ASR systems. However, this requires large amounts of supervised data \cite{radford2022robust}, which is often unavailable in the healthcare domain due to privacy and ethical concerns. 
Consequently, clinical summarization systems often treat ASR systems as black boxes, mandating that overall improvements must arise either from post-processing ASR outputs, or making down-stream methods robust to ASR errors. Recent works propose post-processing ASR outputs using LLMs to correct erroneous ASR transcripts \cite{radhakrishnan2023whispering, bai2024seed}. Nonetheless, based on prior studies, using prompting techniques to clear noise only proves effective only when using LLMs of large sizes, typically those that exceed 100B parameters \cite{yang2023generative}. Moreover, LLMs are prone to hallucinations~\cite{maynez2020faithfulness,nagar2024umedsum}, which can introduce irrelevant symptoms or medication names into the dialogue transcript, potentially degrading summarization quality while attempting to clear noise.

Alternatively, the summarization robustness to ASR errors can be improved through data augmentation techniques \cite{fabbri2021improving}. However, the conventional augmentation approach of exposing the summarization model to erroneous ASR dialogues during the training phase is not feasible either, again due to the limited availability of medical dialogue audio preventing the generation of ASR dialogue transcripts~\cite{nanayakkara2022clinical}. Heuristic approaches for augmentation, such as randomly applying a set of corruption operations, are not ideal either, as they fail to accurately mimic ASR errors both qualitatively and quantitatively, causing the augmented training data distribution to diverge from the real test distribution \cite{wang2020data}.

Recognizing these limitations, we propose the use of LLMs to generate synthetic dialogues mimicking \emph{real} ASR transcriptions with their characteristic errors, as a means for data augmentation. To circumvent hallucinations, we rely on the signal from downstream tasks during fine-tuning on these generated transcripts. As such, if new domain-specific entities are mistakenly hallucinated during the augmentation process, the fine-tuning phase ensures that the model learns to disregard these irrelevant entities, increasing the feasibility of this approach as a result. To accommodate for the scarcity of medical audio recordings, we leverage the in-context learning \cite{brown2020language} capabilities to specialize LLMs for the task of ASR transcript generation based on a few descriptive examples. Specifically, using Primock57 dataset \cite{papadopoulos-korfiatis-etal-2022-primock57}, which includes audio recordings of clinical visits alongside their corresponding human-transcribed text, we first produce noisy transcriptions using ASR models. These noisy ASR dialogue transcripts are then paired with their clean human-transcribed versions to form the few-shot examples needed for effective in-context learning.

While in-context learning enables the qualitative approximation of ASR errors such as phonetic confusions, ensuring that synthetic errors are quantitatively similar remains a challenge \cite{everson2024towards}. To address this, we first analyze the error profile of ASR models by measuring their word-error-rates and the distribution of error types, including insertions, deletions, and substitutions. Subsequently, we introduce a novel description syntax that instructs the LLMs on where to make realistic errors and what types of errors to introduce. By tagging the inputs with appropriate error tags based on the measured noise profiles of the target ASR models, we ensure that the synthetic errors generated are both qualitatively and quantitatively similar to real-world ASR errors. This methodology allows us to create synthetic noisy dialogues that accurately reflect ASR error patterns, which can be used for effective data augmentation in training robust summarization models.

Our experimental evaluation reveals that large language models (LLMs) are effective noise modelers, capable of producing similar errors  to those produced by ASR. This is evident from the qunatitative and qualitative similarities between the word error profiles of the synthetic dialogues we generate and actual ASR dialogue transcripts. Such similarity also reflects on the downstream summarization performance, mirroring the performance drop caused by ASR errors. Lastly, when we utilize these synthetic noisy dialogues to augment the training set of the summarization models, the performance on the noisy test set improves by up to $16\%$, indicating enhanced robustness against ASR errors. In summary, our contributions leading to the robustness are:

\begin{itemize}
    \item We utilize the in-context learning ability of LLMs to create synthetic errors that closely resemble those present in ASR transcriptions. This method is used as a data augmentation strategy to improve the strength of dialogue summarization models in situations where audio recordings are limited.
    \item We introduce an error tagging syntax designed to accurately match the error patterns of real ASR transcriptions in the generated synthetic dialogues. This syntax provides detailed control over the type and amount of errors added, ensuring that the synthetic data closely aligns with real-world ASR errors.
\end{itemize}

\section{Related Work}

\paragraph{Dialogue Summarization}
Summarizing human conversations or speech has been a longstanding challenge that has received significant attention over the years \cite{hori2003new, liu2011speech}. An important subset of these endeavors involves ASR-based summarization, where automatic speech recognition systems are used to transcribe spoken dialogues before summarization \cite{zhong2022dialoglm}. Recently, this challenge has extended to clinical applications, including the summarization of patient-doctor conversations \cite{krishna-etal-2021-generating}, utilizing few-shot~\cite{le2024real} or fine-tuned~\cite{Schlegel2023} LLMs as tools for text comprehension.

\begin{figure*}[t!] 
    \centering
    \subfloat[Pre-processing stage. (i) Human-annotated clean dialogue transcripts are paired with ASR counterparts to compose in-context examples. (ii) Distribution of word-errors is inferred.]{\includegraphics[width=0.5\textwidth]{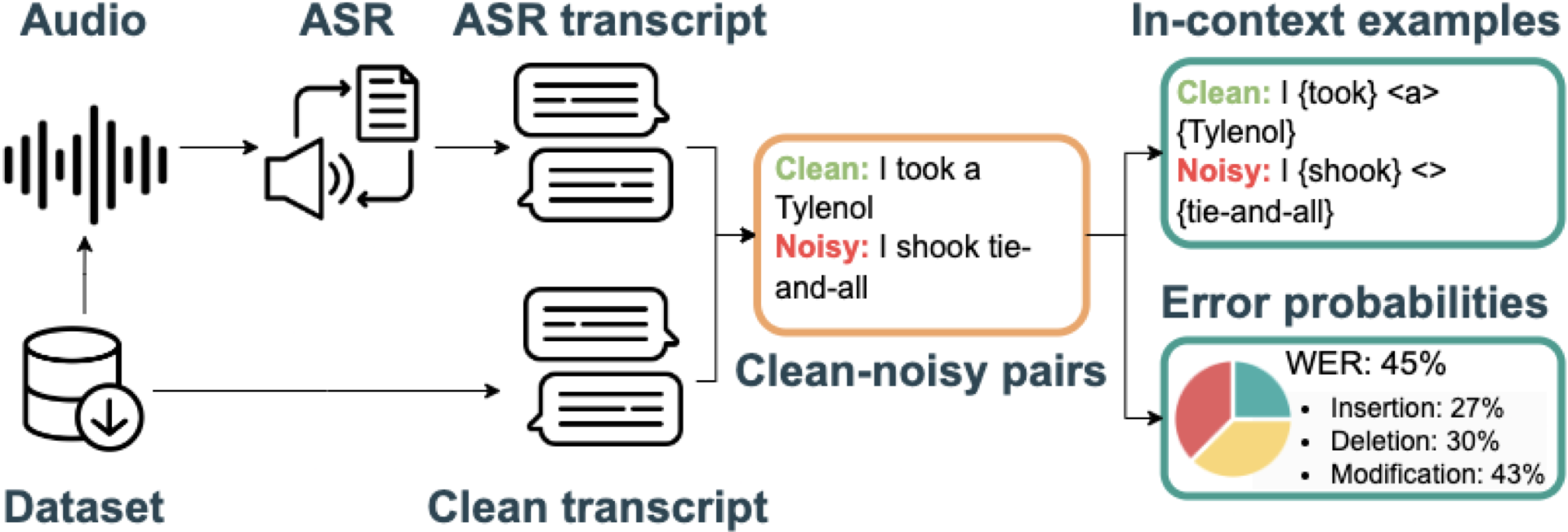}}
    \hspace{3pt}
    \rulesep
    \hspace{3pt}
    \subfloat[Noisy data generation. Inferred word-error distribution is used to inject error tags to the input sentence. Tagged input, instruction prompt and in-context examples are passed to the noise generator.]{\includegraphics[width=0.45\textwidth]{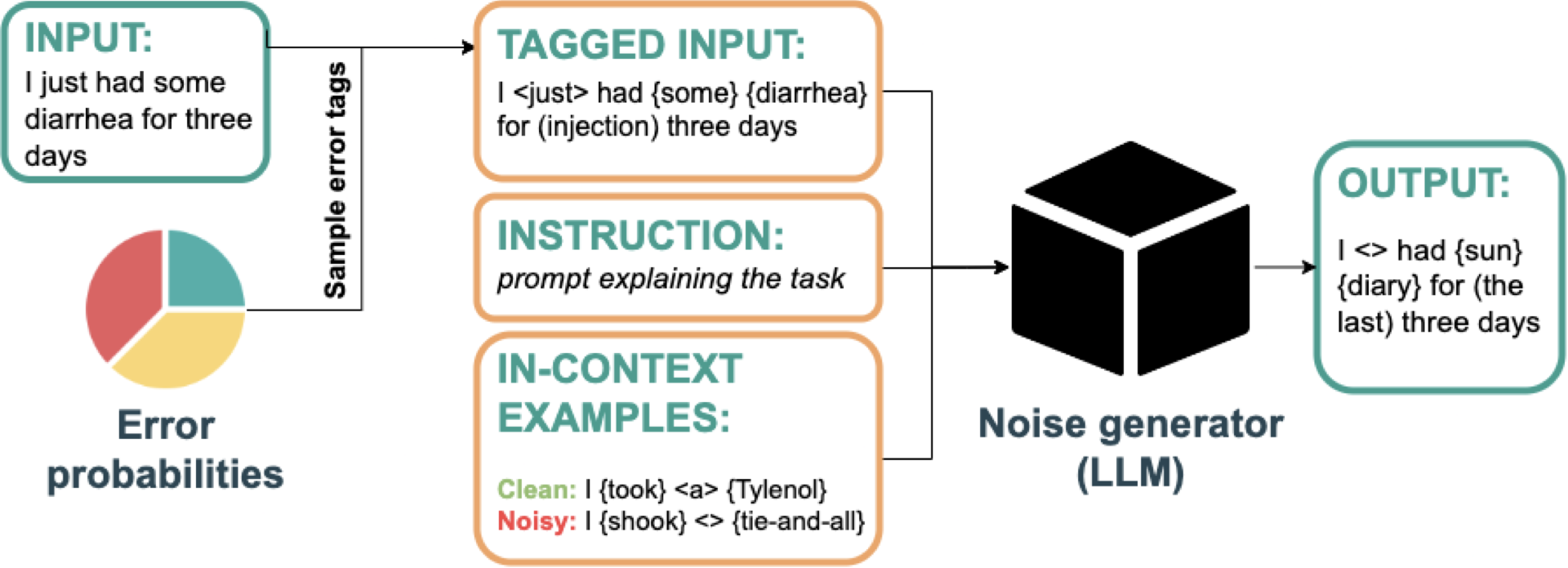}}     
    \caption{Overview of our \frameworkname{} pipeline. First in-context examples are constructed and the error profile of the target ASR model is inferred. Later, the error profile, in-context examples and inputs dialogues are processed by the LLM model to generate noisy dialogues.}
    \label{fig:architecture}
\end{figure*}

\paragraph{ASR Error Correction}
ASR models are prone to producing erroneous dialogue transcripts, especially in challenging environments where the conversation is disrupted by noise. These errors can significantly degrade the quality of downstream tasks like summarization using ASR transcriptions. Recent advancements in LLMs have shown promise in addressing these ASR errors. For instance, \citet{yang2023generative} investigated prompting techniques in correcting ASR errors. \citet{radhakrishnan2023whispering} introduced Whispering LLaMA, a cross-modal generative error correction framework that fuses acoustic information with linguistic representations. \citet{bai2024seed} proposed a seed-based method that enhances ASR outputs by integrating linguistic context during correction, effectively reducing error propagation in downstream tasks. \citet{hu2024large} introduced a method that leverages the inherent noise in audio signals to generate robust language embeddings. To provide a standard benchmark for evaluating ASR error correction methods, \citet{chen2024hyporadise} introduced the Hyporadise dataset. These techniques are unsuitable for the medical domain because fine-tuning denoising models is often impractical due to the lack of public medical dialogue audio recordings \cite{nanayakkara2022clinical}. Additionally, prompting without fine-tuning is only effective for large models \cite{yang2023generative}, which might be impractical, due to privacy concerns when using public APIs or resource constraints when hosting locally.

\paragraph{Data Augmentation}
Data augmentation (DA) is crucial for enhancing model performance by diversifying training examples. Traditional DA methods in NLP, such as paraphrasing~\cite{sharma2023and,li2023team}, back-translation~\cite{sugiyama2019data}, and noise injection~\cite{wang2018switchout}, have been effective in some scenarios, but may not fully capture the complexity of real-world variations. Recent LLM advancements have revolutionized DA by using models like GPT-3 and GPT-4 to generate high-quality synthetic data. For instance, \citet{chintagunta-etal-2021-medically} use in-context learning with LLMs to create synthetic medical dialogue summaries, while Dialogic~\cite{li2022controllable} employs LLMs to generate annotated dialogues with automatic verification and revision. Similarly, \citet{liu2024context} leverage topic-focused summarization and domain adaptation with LLMs to generate personalized medical dialogues. DA is also used to improve robustness against adversarial examples, as used by~\citet{liu2020adversarial}, who apply adversarial training to make sentiment analysis models more resilient to noise and variations.
Our method differs by addressing the challenges of augmenting medical dialogue data where real ASR transcripts are scarce. Instead of relying on heuristic corruptions that don't mimic ASR errors accurately, we use LLMs to generate synthetic dialogues that realistically replicate ASR errors, ensuring the augmented training data better aligns with real-world test conditions.

\section{MEDSAGE}
We use LLMs to generate realistic ASR noise. On a high level, we pair human-transcribed sentences with their ASR counterparts as in-context examples to an LLM, which we instruct to generate noisy sentences (Section~\ref{sec:err-gen}). These examples convey the qualitative aspects of the ASR errors such as phonetic confusions. To also ensure that the errors in the generated synthetic dialogues quantitatively match that of the ASR transcriptions, we propose a controlled generation strategy (Section~\ref{sec:ctrl-gen}). Our strategy involves introducing a tagging syntax that instructs the LLM on the specific word locations to corrupt with specific types of errors. This enables the generation of synthetic dialogues with any arbitrary error distribution that mirrors the characteristics observed in real ASR transcriptions. An overview of our \frameworkname{} pipeline is given in Figure \ref{fig:architecture}.

\subsection{Error Generation using In-context Learning}
\label{sec:err-gen}
To generate synthetic noisy dialogues, we first use a specialised system prompt to instruct the LLMs about the error generation task. Later we form the in-context examples by pairing sentences from human-transcribed clean dialogues with their ASR-generated counterparts. These examples are subsequently provided to the LLMs to convey the characteristics of ASR errors that we are trying to mimic. An example query that illustrates our structure when prompting the noise-generating LLMs is displayed in the following box. 
\begin{tcolorbox}[colback=white, colframe=black, title=In-Context Learning for ASR Noise Generation, bottom=0pt]
\noindent \textbf{\#\#\# System Prompt:} \\
You are an AI assistant tasked with simulating errors similar to those made by Automated Speech Recognition (ASR) systems. You will be given sentences with the type of errors to be made indicated by tags. Corrupt the sentences based on \texttt{[explanation of the tagging system]}.\\
\\
\begin{tcolorbox}[colback=white, colframe=black, sharp corners, title=In-Context Examples, boxsep=0pt, before=\vspace{0pt}, after=\vspace{10pt}]
\noindent \textbf{\#\#\# Input:} I \textbf{\{}took a Tylenol\textbf{\}} \texttt{(human-} \\
\texttt{transcribed)}\\
\noindent \textbf{\#\#\# Response:} I \textbf{\{}shook tie-and-all\textbf{\}} \texttt{(ASR)}\\
\\
\noindent \textbf{\#\#\# Input:} I just had some \textbf{\{}diarrhea\textbf{\}} for the last three days \texttt{(human-transcribed)}\\
\noindent \textbf{\#\#\# Response:} I just had some \textbf{\{}diary\textbf{\}} for the last three days  \texttt{(ASR)}\\
\end{tcolorbox}
\noindent \textbf{\#\#\# Input:} yeah now i mean have you any have you noticed any kind of \textbf{\{}white spots\textbf{\}} on the back of your back of your \textbf{\{}throat\textbf{\}} or redness \\
\noindent \textbf{\#\#\# Response:}  \\
\end{tcolorbox}
\par
As seen from the example, the clean sentences are given as inputs and the responses are expected to contain erroneous versions as if they were obtained through ASR models. The curly braces surrounding certain substrings are a part of our error tagging system, which is detailed later in Section \ref{sec:ctrl-gen}, and they denote which words or substrings are transcribed wrongly by the ASR models and how. For instance, in the example provided, the word ``Tylenol'' is wrongly transcribed as ``tie-and-all'', highlighting the common issue of ASR systems confusing tokens with similar pronunciation. Lastly, the input at the end contains the actual clean sentence that we aim to corrupt. The words to be corrupted are also indicated to the LLM through the use of curly braces.
\subsection{Controlled Generation}
\label{sec:ctrl-gen}
For data augmentation to be effective in improving performance on noisy test data, the generated synthetic data must closely follow the error patterns found in real-world ASR outputs. However, solely relying on in-context learning with a few examples often fails to capture the nuanced distribution of errors produced by ASR models. To overcome this limitation, we propose a controlled noise injection mechanism. This is to, ensuring the synthetic data aligns with the error profile of the target ASR model.
This approach relies on a detailed analysis of the types and frequencies of errors in ASR transcripts. Using these insights, we then guide the noise generation process by conditioning the LLM with an error tagging syntax.

\paragraph{ASR Error Profiling:} 
To quantitatively represent the error profile of ASR models, we focus on three primary error types: insertion, deletion, and substitution, which collectively contribute to the calculation of Word Error Rate (WER). To estimate the distribution of these errors, we first transcribe a set of medical conversation dialogues using the ASR models. The transcriptions are then aligned with human-annotated ground truth using the Wagner-Fisher algorithm \cite{wagner1974string}. This alignment allows us to and quantify the occurrences of each specific error type produced by the ASR model.

We estimate the probability distribution of errors as follows. Let  $c_i$  denote the event that the word at index  $i$  is corrupted. The probability  $p(c_i)$  of a word being corrupted is defined as the WER of the ASR model i.e.,  $p(c_i)$ = WER. Given that a word is corrupted, the probability of a specific error type  $e_t$  can be expressed as  $p(e_t | c_i)$, where  $e_t \in \{\textit{insertion, deletion, substitution}\}$ . These conditional probabilities represent the distribution of different error types observed in the ASR model’s output (See Section \ref{sec:error-profiling}). We use this error distribution to formulate our tagging system.

\paragraph{Tagging System}
We develop a tagging system to instruct the LLM on the specific word-level corruptions to perform to be able to have more control on the WER of generated noise and the distribution of error types. Specifically, we employ the following tags to indicate the error-type that the model should simulate at the word level.

\begin{itemize}
\item Words enclosed in curly brackets \texttt{\{\}} should be replaced with phonetically similar words. For instance, \texttt{\{wheezy\}} might be replaced with \texttt{\{weesy\}}, and \texttt{\{Tylenol\}} could be changed to \texttt{\{tie-and-all\}}.
\item The tag \texttt{(INSERTION)} indicates where a new word should be inserted. These new words should be general in nature and should not introduce new domain-specific terminology such as drug names or symptoms.
\item We do not specify any tags for deletion; instead, words that need to be deleted are simply removed from the text. 
\end{itemize}
\noindent
This meaning of the tagging system is conveyed to the the model both through the system prompt and in-context examples. To decorate the in-context examples with error tags accordingly, we align the clean and noisy examples pairs using the Wagner-Fisher algorithm and determine the locations of word errors along with their types. During inference, these tags are randomly applied on the ground-truth transcripts based on the estimated error distribution of the target ASR model. The probability of tagging a word at index $i$ with a specific error type is thus given by:
\begin{equation}
p(e_t | c_i) \cdot p(c_i)
\end{equation}
which is the joint probability of a word being corrupted and the occurrence of a specific error type.

\begin{algorithm}[!t]
\caption{Generating Synthetic Noisy Dialogues Using LLMs}
\label{alg:noisy_dialogues}
\begin{algorithmic}[1]
\State \textbf{Input:} Clean transcripts \( T_{\text{cln}} \), in-context example pairs \( E_{\text{in}} \), LLM
\State \textbf{Output:} Synthetic noisy dialogues \( T_{\text{syn}} \)
\State
\State \# Create in-context examples
\For{each pair \( (t_{\text{cln}}, t_{\text{ASR}}) \) in \( E_{\text{in}} \)}
    \State Compute word-locations and types of errors \( e_i \), \( e_t \)
    \State \( (t_{\text{cln}}, t_{\text{ASR}})' \gets \text{insert\_tags}(t_{\text{cln}}, t_{\text{ASR}}, e_i, e_t) \)
    \State Replace \( (t_{\text{cln}}, t_{\text{ASR}}) \) with \( (t_{\text{cln}}, t_{\text{ASR}})' \) in \( E_{\text{in}} \)
\EndFor
\State
\State Initialize \( T_{\text{syn}} \gets \emptyset \)
\State
\State \# Prompt LLM with in-context examples
\For{each input dialogue \( t_{\text{cln}} \) in \( T_{\text{cln}} \)}
    \State Sample error indexes \( e_i \sim p(c_i) \)
    \State Sample error types \( e_t \sim p(e_t \mid c_i = e_i) \)
    \State \( t_{\text{cln}}' \gets \text{insert\_tags}(t_{\text{cln}}, e_i, e_t) \)
    \State \( t_{\text{syn}} \gets \text{LLM}(\text{prompt}, t_{\text{cln}}') \)
    \State \( T_{\text{syn}} \gets T_{\text{syn}} \cup t_{\text{syn}} \)
\EndFor
\end{algorithmic}
\end{algorithm}

\section{Experiment Settings}
This section presents our experimental evaluation of the proposed method for generating synthetic noisy dialogue transcripts to improve the robustness of summarization models against ASR errors. 

\paragraph{ASR models: } We used the \textit{Whisper tiny/ large}\cite{radford2023robust}, and \textit{Wav2vec2-base} \cite{baevski2020wav2vec} ASR models to generate transcriptions of medical dialogues. \paragraph{Large Language Models: } We use \textit{Llama-3-8B} \cite{dubey2024llama} and \textit{Mistral-7B} \cite{jiang2023mistral} both for generating synthetic dialogues and summarization, while \textit{Gemma-7B} \cite{team2024gemma} is only used for summarization.
\paragraph{Datasets: } For testing our approach, we use the \textit{Primock57} dataset, which comprises audio recordings of 57 enacted doctor-patient dialogues and their text transcripts written by human annotators. For experiments involving fine-tuning we use the \textit{NoteChat-1000} dataset \cite{wang2024notechat} for training, which includes 1000 synthetic doctor-patient dialogue transcripts generated by multiple LLMs in a cooperative roleplay setting, conditioned on clinical notes.

\paragraph{Evaluation Metrics:}
We utilized three types of evaluation metrics to assess the performance of the summarization models. For lexical similarity, we used the ROUGE metrics \cite{lin2004rouge}, specifically focusing on ROUGE-L, which measures the longest common subsequence overlap between generated and reference summaries. To capture the semantic similarity, we employed BERTScore \cite{zhangbertscore}, which uses embeddings from pre-trained BERT models to compare the contextual meaning of the texts. Additionally, recognizing the importance of accurately identifying medical terminology in summaries, we assessed the overlap of domain-specific named entities (DSEs) using the F1 score, which combines both precision and recall, based on entities extracted through named entity recognition (NER).

\section{Preliminary Experiments}
In this section, we present preliminary experiments that lay the foundation for our work. First, we conduct a motivating study that verifies how ASR errors impair the performance of a downstream medical dialogue summarization task. Subsequently, we perform an analysis that suggests different ASR models exhibit distinct error profiles.

\begin{table}[!b]
\centering
\label{tab:summarization_with_denoising}
\resizebox{\columnwidth}{!}{%
\begin{tabular}{lcccccc}
\toprule
 & \multicolumn{3}{c}{\textbf{Llama-3-8B}} & \multicolumn{3}{c}{\textbf{Mistral-7B}} \\
\cmidrule(lr){2-4} \cmidrule(lr){5-7}
\textbf{Transcription} & \textbf{F1} & \textbf{RougeL} & \textbf{Bert} & \textbf{F1} & \textbf{RougeL} & \textbf{Bert} \\
\midrule
\textbf{Wav2vec2-base (ASR)} & 16.99 & 11.52 & 52.46 & 17.06 & 11.09 & 50.40 \\
\hspace{0.3cm}\textbf{+ Denoising} & 15.20 & 10.19 & 51.49 & 20.62 & 11.03 & 51.51 \\
\textbf{Whisper-tiny (ASR)} & 19.86 & 11.65 & 52.73 & 21.85 & 11.91 & 51.97 \\
 \hspace{0.3cm}\textbf{+ Denoising} & 12.73 & 9.95 & 51.94 & 19.81 & 11.42 & 51.74 \\
 \textbf{Whisper-large (ASR)} & 21.42 & 13.67 & \textbf{54.23} &  24.06 & \textbf{13.27} & \textbf{53.29} \\
\hspace{0.3cm}\textbf{+ Denoising} & 18.64 & 12.57 & 53.15 & 23.12 & 12.67 & 52.50 \\
\midrule
\textbf{Ground Truth} & \textbf{22.11} & \textbf{14.07} & 53.80 & \textbf{24.53} & 13.25 & 52.85 \\
\bottomrule
\end{tabular}%
}
\caption{Effect on Summarization quality due to ASR errors. \textit{``Ground Truth''} indicates using the ground-truth ASR transcripts as input for summarization for measuring upper-bound performance. \textit{+ Denoising} indicates that the ASR generated transcripts are passed through a denoising LLM to clean the errors if any and then summarized.}
\label{tab:impact-on-summarization}
\end{table}
\subsection{ASR noise harms medical report quality}
\label{subsec:impact_asr_noise}

Our findings, as displayed in Table \ref{tab:impact-on-summarization}, reveal that ASR noise can significantly degrade the quality of the generated summaries, especially when using small ASR models. Specifically, using Wav2vec2-base generated transcripts instead of human annotated ones lead to a noticeable reduction in the domain-specific entity F1 score of $23\%$ ($22.11 \rightarrow 16.99$). While using larger models like Whisper-large exhibits comparable downstream task performance as using ground truth dialogues, deployment area of such large-scale models is limited due to the computational resource requirements they demand. Such a difference in the DSE overlap measure indicates a loss in accurately capturing critical medical entities. The ROUGE-L scores also decline, reflecting reduced textual overlap and coherence between the generated and reference summaries. Moreover, the BERT Scores drop, suggesting a decrease in the semantic similarity to the reference summaries. These results underscore the challenges of downstream tasks, such as dialogue summarisation, face, arising from erroneous ASR transcripts. We also explored applying few-shot denoising on ASR transcribed dialogues, which is denoted as ``+ Denoising''. However, the results show that this method does not recover the summarization performance, highlighting the limitations of traditional post-processing based noise reduction techniques in this context.

\subsection{Different ASR models exhibit different error profiles}
\label{sec:error-profiling}
\begin{figure}[!ht]
    \centering
    \includegraphics[width=0.85\columnwidth]{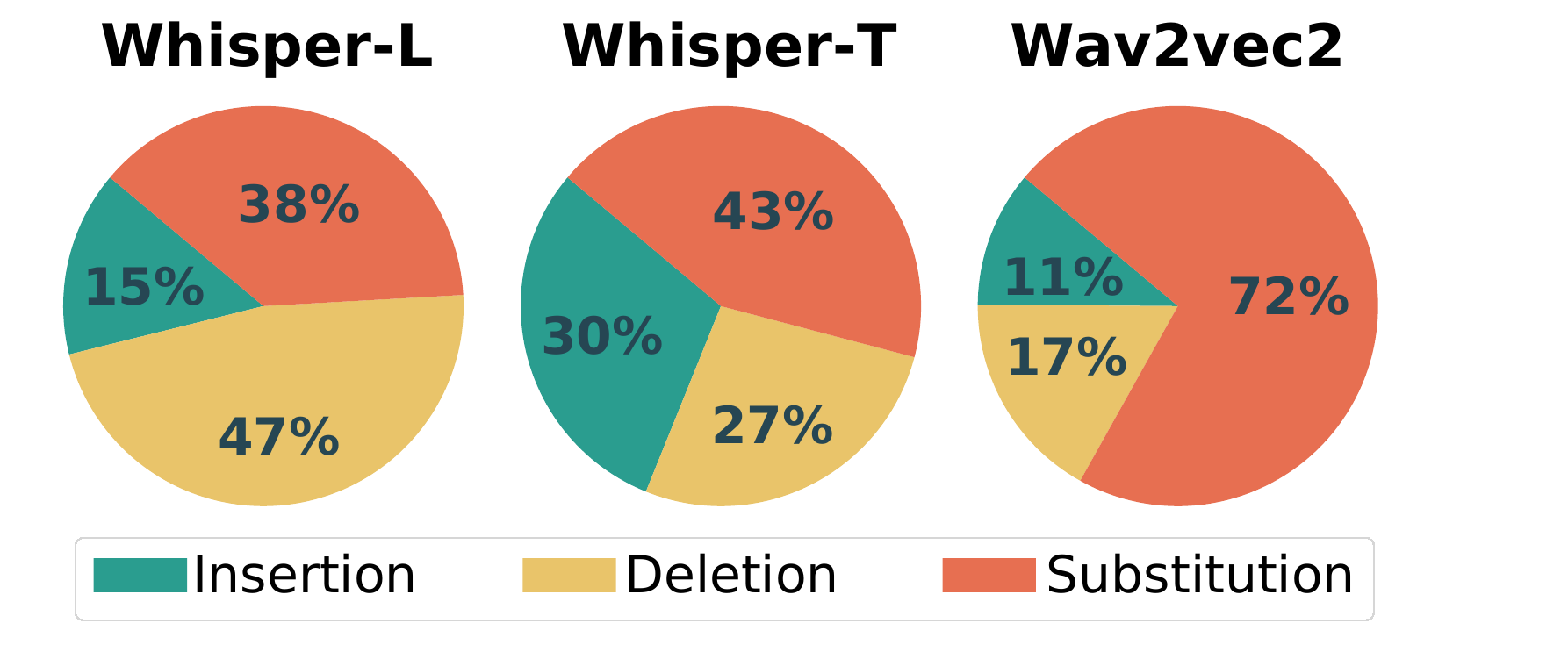}
    \caption{ASR errors of different models. The breakdown shows the different types of errors made by the model.}
    \label{fig:error_profile}
\end{figure}

We analyzed the errors made by the  ASR models on Primock57 audio samples. \textit{Whisper-large} transcription results in $25\%$ WER. Both the \textit{Whisper-tiny} and \textit{Wav2vec2-base} models had similar WER scores, with the former achieving $44\%$ and the latter $45\%$. Moreover, the breakdown of error types associated with each ASR model is displayed in Figure \ref{fig:error_profile}. The differing noise profiles suggest that to accurately mimic the properties of ASR transcriptions, the synthetic dialogue generation process must be controllable and adjustable with respect to the error profile of the target ASR model.

\section{Main Experiments}
Building on the insights from our preliminary experiments, we present our main experimental findings in this section.

\begin{table*}[!t]
\centering
\resizebox{\textwidth}{!}{%
\begin{tabular}{lccccccccc}
\toprule
 & \multicolumn{3}{c}{\textbf{Llama-3-8B}} & \multicolumn{3}{c}{\textbf{Mistral-7B}} & \multicolumn{3}{c}{\textbf{Gemma-7B}} \\
\cmidrule(lr){2-4} \cmidrule(lr){5-7} \cmidrule(lr){8-10} 
\textbf{Method} & \textbf{F1} & \textbf{RougeL} & \textbf{Bert} & \textbf{F1} & \textbf{RougeL} & \textbf{Bert} & \textbf{F1} & \textbf{RougeL} & \textbf{Bert}  \\
\midrule
\textbf{Zero-shot} & 19.90 & 11.84 & \textbf{52.85} & 20.77 & 11.81 & 52.36 & 21.30 & 12.73 & 50.95 \\
\hspace{0.3cm}\textbf{+ Denoising} & 15.31 \scriptsize{(-23.0\%)} & 10.55 \scriptsize{(-10.9\%)} & 51.93 \scriptsize{(-1.7\%)} & 17.87 \scriptsize{(-14.0\%)} & 10.75 \scriptsize{(-9.0\%)} & 51.32 \scriptsize{(-2.0\%)} & 18.97 \scriptsize{(-10.9\%)} & 12.63 \scriptsize{(-0.8\%)} & 51.11 \scriptsize{(+0.3\%)} \\
\textbf{FT on clean} & 20.42 \scriptsize{(+2.6\%)} & 12.74 \scriptsize{(+7.6\%)} & 52.35 \scriptsize{(-1.0\%)} & 16.76 \scriptsize{(-19.3\%)} & 8.06 \scriptsize{(-31.7\%)} & 49.46 \scriptsize{(-5.5\%)} & 21.21 \scriptsize{(-0.4\%)} & 13.19 \scriptsize{(+3.6\%)} & 51.31 \scriptsize{(+0.7\%)} \\
\hspace{0.3cm}\textbf{+ Denoising} & 19.34 \scriptsize{(-2.8\%)} & 11.49 \scriptsize{(-3.0\%)} & 51.35 \scriptsize{(-2.8\%)} & 14.88 \scriptsize{(-28.3\%)} & 7.49 \scriptsize{(-36.6\%)} & 48.73 \scriptsize{(-6.9\%)} & 19.09 \scriptsize{(-10.4\%)} & 12.49 \scriptsize{(-1.9\%)} & 50.57 \scriptsize{(-0.7\%)} \\
\midrule
\textbf{FT-aug (1x, \frameworkname)} & 20.93 \scriptsize{(+5.2\%)} & 12.29 \scriptsize{(+3.8\%)} & 52.00 \scriptsize{(-1.6\%)} & 23.89 \scriptsize{(+15.0\%)} & 12.34 \scriptsize{(+4.5\%)} & 52.68 \scriptsize{(+0.6\%)} & 22.84 \scriptsize{(+7.2\%)} & 13.33 \scriptsize{(+4.7\%)} & 51.42 \scriptsize{(+0.9\%)} \\
\textbf{FT-aug (2x, \frameworkname)} & 22.84 \scriptsize{(+14.8\%)} & 13.08 \scriptsize{(+10.5\%)} & 51.36 \scriptsize{(-2.8\%)} & 24.17 \scriptsize{(+16.4\%)} & 13.27 \scriptsize{(+12.4\%)} & 53.10 \scriptsize{(+1.4\%)} & 22.48 \scriptsize{(+5.5\%)} & 12.73 \scriptsize{(0.0\%)} & 51.48 \scriptsize{(+1.0\%)} \\
\textbf{FT-aug (3x, \frameworkname)} & 22.51 \scriptsize{(+13.1\%)} & 12.78 \scriptsize{(+7.9\%)} & 51.91 \scriptsize{(-1.8\%)} & 22.11 \scriptsize{(+6.5\%)} & 12.93 \scriptsize{(+9.5\%)} & 52.66 \scriptsize{(+0.6\%)} & 20.96 \scriptsize{(-1.6\%)} & 13.25 \scriptsize{(+4.1\%)} & 51.29 \scriptsize{(+0.7\%)} \\
\midrule
\textbf{\textbf{FT-aug (Best, \frameworkname)}} & \textbf{22.84 \scriptsize{(+14.8\%)}} & \textbf{13.08 \scriptsize{(+10.5\%)}} & 52.00 \scriptsize{(-1.6\%)} & \textbf{24.17 \scriptsize{(+16.4\%)}} & \textbf{13.27 \scriptsize{(+12.4\%)}} & \textbf{53.10 \scriptsize{(+1.4\%)}} & \textbf{22.84 \scriptsize{(+7.2\%)}} & \textbf{13.33 \scriptsize{(+4.7\%)}} & \textbf{51.48 \scriptsize{(+1.0\%)}} \\
\bottomrule
\end{tabular}%
}
\caption{Summarization qualities of Llama-3-8B, Mistral-7B, and Gemma-7B for different training settings. \textit{FT} and \textit{aug} stand for \textit{fine-tuning} and \textit{data augmentation} respectively. The number of synthetic dialogues per ground truth dialogue included in augmented training sets are indicated with $x$ multiples. Relative improvements compared to the Zero-shot method are displayed in parenthesis ($\%$).}
\label{tab:summarization_methods}
\end{table*}
\subsection{Data augmentation using synthetic dialogues improves robustness against ASR errors}
\label{subsec:adversarial_training}
To assess the impact of \frameworkname{}, we first investigate the effectiveness of LLM-generated synthetic dialogues in building robustness against ASR errors through data augmentation.  In this experiment, we begin by augmenting the training set of the note chat-1000 dataset using synthetic dialogues. Later, we fine-tune summarization models through LoRa (Low-Rank Adaptation) adapters \cite{hu2021lora} on augmented datasets that include a mix of clean ground-truth dialogues and our synthetic noisy dialogues. Then we compare the summarization performance of these fine-tuned models against three baselines on the ASR-transcribed Primock57 audio recordings, which comprise our test set. The three baselines we use for evaluation are:
\begin{itemize}
    \item \textbf{Zero-shot}: Pre-trained LLMs are prompted to replicate medical notes written by doctors based on input dialogues.
    \item \textbf{FT on clean}: LLMs are fine-tuned only on clean transcripts.
    \item \textbf{Denoising}: The ASR transcripts are cleaned by LLama-3-8B model before summarization.
\end{itemize}
As shown in Table~\ref{tab:summarization_methods}, the results indicate that the inclusion of synthetic noisy dialogues in the training set considerably improves the robustness of the summarization models, resulting in up to $16.4\%$ improvement in F1. In contrast, \textit{FT on clean} baseline only marginally improves or even degrades performance depending on the summarization model architecture. This suggests that while fine-tuning on clean dialogues alone can benefit summarization models by adapting them to the medical domain, it does not consistently enhance their ability to handle ASR noise. Moreover, denoising consistently harmed summarization performance across all experiments. This can be attributed to the two aforementioned key factors: (1) denoising models below a certain parameter size (such as LLama-8B) are ineffective at cleaning noise, and (2) using prompting techniques without fine-tuning introduces a risk of hallucinations, potentially injecting irrelevant medical entities into the dialogue.
\subsection{LLM-produced synthetic errors are realistic}
\label{subsec:synthetic_noisy_transcripts}
To ensure that the improvements observed indeed stem from the accurate simulation of ASR noise, we need to establish whether the synthetic dialogues generated by our \frameworkname{} method exhibit error characteristics similar to those found in actual ASR transcriptions. This assessment involves two main comparisons.
\begin{figure}[!ht]
\centering
\resizebox{0.9\columnwidth}{!}{%
\begin{tabular}{p{1.20\columnwidth}}
\hline
 \textbf{Sentence} \\
\hline
 \textbf{(Ground truth)} Doctor: yeah now i mean have you any \yellowhl{have} you \yellowhl{noticed} \redhl{any} \redhl{kind} \yellowhl{of} white spots on the back \yellowhl{of} \yellowhl{your} back of your \yellowhl{throat} or \yellowhl{redness} \\
 \textbf{(Whisper ASR)} Doctor: yeah now i mean have you any \yellowhl{if} you \yellowhl{know} \redhl{the} \redhl{new} \yellowhl{chrome} white spots on the back \yellowhl{ports} \yellowhl{youll} back of your \yellowhl{throws} or \yellowhl{readiness} \\
\hline
\textbf{(Ground truth)} Doctor: yeah now i mean have you any \yellowhl{have} \greenhl{you} \yellowhl{noticed} any kind \yellowhl{of} \greenhl{white} \greenhl{spots} on the back of your \greenhl{back} \yellowhl{of} \yellowhl{your} \yellowhl{throat} or \yellowhl{redness} \\
\textbf{{\texttt{MEDSAGE}}} Doctor: yeah now i mean have you any \yellowhl{do} \greenhl{ya} \yellowhl{notice} any kind \yellowhl{kinda} \greenhl{whish} \greenhl{spits} on the back of your \greenhl{bak} \yellowhl{o} \yellowhl{yer} \yellowhl{throt} or \yellowhl{reddness} \\
\hline
\end{tabular}
}
\caption{Comparison of an ASR transcription and its LLM-generated counterpart produced by \texttt{MEDSAGE}. The words highlighted in yellow are errors at common word indexes among both transcripts, while those that are highlighted in red and green indicate unique errors of ASR and \texttt{MEDSAGE}.}
\label{fig:qualitative-comparison}
\end{figure}
\begin{figure}[!b]
    \centering
    \includegraphics[width=0.95\columnwidth]{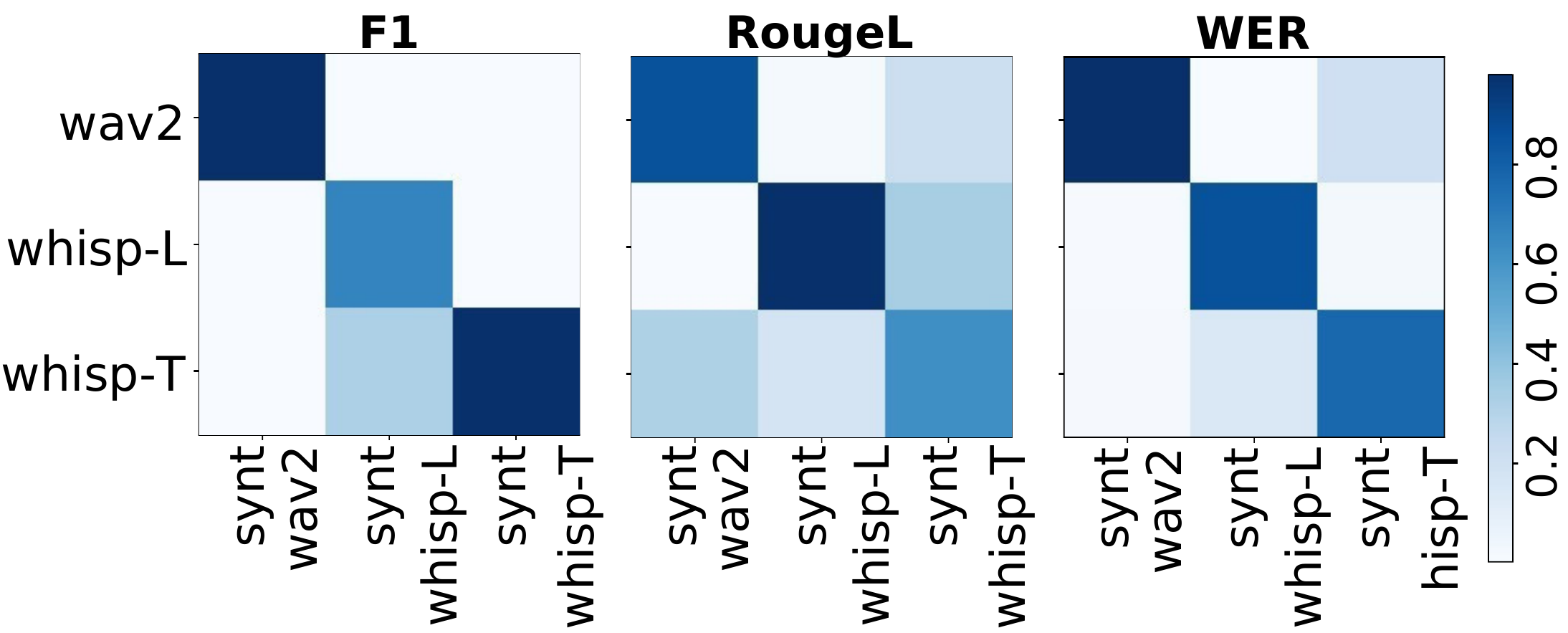}
    \caption{Similarities between ASR-generated transcriptions (rows) and LLM-generated synthetic transcriptions (columns) with respect to F1, Rouge-L, and WER metrics. The highest similarities are observed on the diagonals indicating overlap between corresponding ASR- and LLM-generated transcriptions.}
    \label{fig:similarity-matrices}
\end{figure}
\subsubsection{Qualitative Comparison}
To illustrate that synthetic dialogues produced by our \frameworkname{} method properly mimics the errors characteristics that real ASR transcriptions exhibit, we present randomly selected snippets from doctor-patient conversations in Figure~\ref{fig:qualitative-comparison}. 
The commonalities between the nature of errors present in both utterances, such as phonetic confusions underscore the accuracy of the noise injections. For instance, the words "white spots" are confused with "whish spits". This qualitative evidence supports our quantitative findings, demonstrating that LLM-generated synthetic noise can replicate the nuanced flaws typically seen in ASR outputs.

\begin{table}[!t]
\centering
\resizebox{0.8\columnwidth}{!}{%
\begin{tabular}{lcccc}
\toprule
 & \multicolumn{4}{c}{\textbf{Llama-3-8B Summarization}} \\
\cmidrule(lr){2-5}
\textbf{Source Transcripts} & \textbf{\textcolor{blue}{ASR}} & \textbf{Random} & \multicolumn{2}{c}{\textbf{LLM-generated (\frameworkname{})}} \\
\cmidrule(lr){2-2} \cmidrule(lr){3-3} \cmidrule(lr){4-5}
 & & & \textbf{Llama-3-8B} & \textbf{Mistral-7B} \\
\midrule
\multicolumn{5}{c}{\cellcolor[HTML]{C0C0C0}\textbf{F1}} \\
\textbf{Wav2vec2-base}  & \textcolor{blue}{16.99} & 11.24 & \textbf{17.35} & 16.54 \\
\textbf{Whisper-tiny}   & \textcolor{blue}{19.86} & 12.83 & \textbf{18.02} & 17.91 \\
\textbf{Whisper-large}   & \textcolor{blue}{21.42} & 18.84 & 20.35 & \textbf{22.02} \\
\midrule
\multicolumn{5}{c}{\cellcolor[HTML]{C0C0C0}\textbf{RougeL}} \\
\textbf{Wav2vec2-base}  & \textcolor{blue}{11.52} & 8.50 & \textbf{11.47} & 11.18 \\
\textbf{Whisper-tiny}   & \textcolor{blue}{11.65} & 9.88 & \textbf{11.92} & 10.91 \\
\textbf{Whisper-large}   & \textcolor{blue}{13.67} & 11.98 & 12.52 & \textbf{12.70} \\
\midrule
\multicolumn{5}{c}{\cellcolor[HTML]{C0C0C0}\textbf{Bert}} \\
\textbf{Wav2vec2-base}  & \textcolor{blue}{52.46} & 49.59 & 52.87 & \textbf{52.71} \\
\textbf{Whisper-tiny}   & \textcolor{blue}{52.73} & 50.80 & \textbf{53.48} & 51.57 \\
\textbf{Whisper-large}   & \textcolor{blue}{54.23} & 52.07 & \textbf{53.70} & 53.59 \\
\bottomrule
\end{tabular}%
}

\vspace{5pt}

\resizebox{0.8\columnwidth}{!}{%
\begin{tabular}{lcccc}
\toprule
 & \multicolumn{4}{c}{\textbf{Mistral-7B Summarization}} \\
\cmidrule(lr){2-5}
\textbf{Source Transcripts} & \textbf{\textcolor{blue}{ASR}} & \textbf{Random} & \multicolumn{2}{c}{\textbf{LLM-generated (\frameworkname{})}} \\
\cmidrule(lr){2-2} \cmidrule(lr){3-3} \cmidrule(lr){4-5}
 & & & \textbf{Llama-3-8B} & \textbf{Mistral-7B} \\
\midrule
\multicolumn{5}{c}{\cellcolor[HTML]{C0C0C0}\textbf{F1}} \\
\textbf{Wav2vec2-base}  & \textcolor{blue}{17.06} & 11.73 & \textbf{15.95} & 15.05 \\
\textbf{Whisper-tiny}   & \textcolor{blue}{18.04} & 15.38 & 18.63 & \textbf{18.10} \\
\textbf{Whisper-large}   & \textcolor{blue}{24.06} & 22.28 & \textbf{23.82} & 21.40 \\
\midrule
\multicolumn{5}{c}{\cellcolor[HTML]{C0C0C0}\textbf{RougeL}} \\
\textbf{Wav2vec2-base}  & \textcolor{blue}{11.09} & 8.32 & \textbf{11.70} & 10.34 \\
\textbf{Whisper-tiny}   & \textcolor{blue}{11.91} & 9.93 & \textbf{12.60} & 10.51 \\
\textbf{Whisper-large}   & \textcolor{blue}{13.27} & 11.82 & \textbf{13.47} & 11.39 \\
\midrule
\multicolumn{5}{c}{\cellcolor[HTML]{C0C0C0}\textbf{Bert}} \\
\textbf{Wav2vec2-base}  & \textcolor{blue}{50.40} & 44.61 & 51.78 & \textbf{49.56} \\
\textbf{Whisper-tiny}   & \textcolor{blue}{51.97} & 48.17 & \textbf{52.47} & 48.67 \\
\textbf{Whisper-large}   & \textcolor{blue}{53.29} & 50.09 & \textbf{53.64} & 49.94 \\
\bottomrule
\end{tabular}%
}
\caption{Summarization qualities for ASR and LLM-generated dialogue transcripts. \textit{Random} stands for augmenting clean dialogues using random insertion, deletion, and substitutions. The ASR columns are colored blue to indicate reference values. The summarization qualities from synthetic dialogues closest to real ASR are shown in bold.}
\label{tab:summarization-similarity}
\end{table}
\subsubsection{Quantitative Comparison}
To further substantiate the realism of LLM-generated synthetic noise, we conduct quantitative comparisons between the errors in synthetic dialogues and those in ASR-transcribed dialogues using the aforementioned metrics. First, we compute pairwise similarities, w.r.t. three evaluation metrics, among ASR transcripts and synthetic dialogues. As shown in the similarity matrices in Figure~\ref{fig:similarity-matrices}, the diagonal entries contain the highest similarity scores, indicating that the greatest overlap is between ASR transcripts and the corresponding synthetic dialogues generated to mimic them. This pattern suggests that the LLM-generated synthetic dialogues are successful in accurately replicating the specific error characteristics of the ASR transcripts they were designed to imitate. Additionally, we provide an indirect comparison based on the summarization qualities each dialogue transcript yields. In this experiment, we also include dialogues corrupted with random deletions as well as insertions and substitutions from the NLTK words corpus \cite{bird2009natural} and name this baseline as \textit{Random}. As shown in Table~\ref{tab:summarization-similarity}, the summarization qualities of LLM-generated synthetic dialogues closely align with those of the ASR transcripts, which are indicated in blue as reference values. This further supports the effectiveness of our method in producing realistic synthetic data.
\subsection{Error tagging system enables controllable generation}
\label{subsec:noise_rate_effect}
\begin{figure}[!t]
    \centering
\includegraphics[width=0.8\columnwidth]{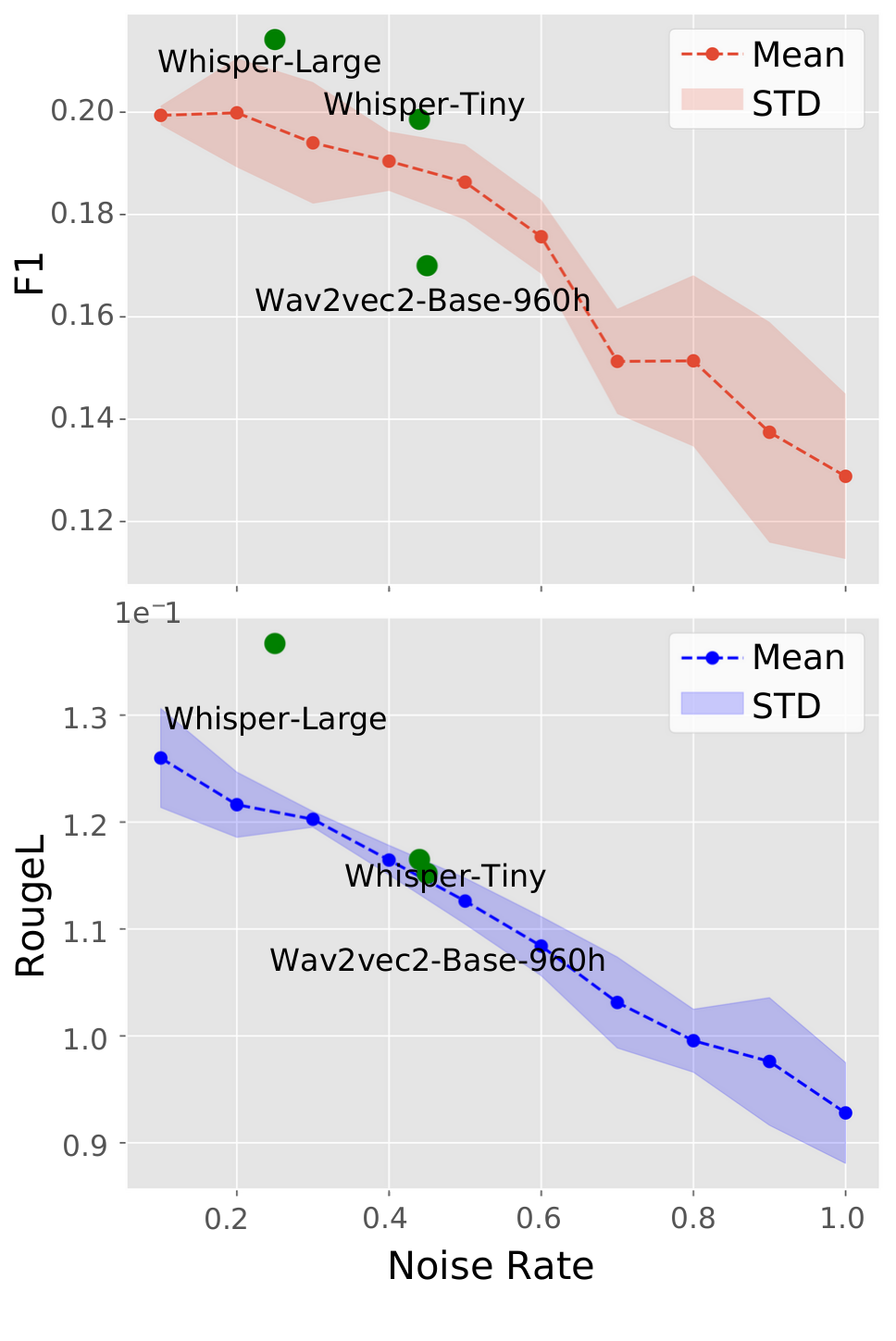}
    \caption{Change in the downstream summarization quality w.r.t. F1 and RougeL metrics as the rate of error tags used to generate synthetic dialogues increases. In-context examples were taken from Whisper-tiny transcribed dialogues. Green dots mark the scores of real ASR transcriptions.}
    \label{fig:noise_vs_quality}
\end{figure}

Lastly, we validate that our method is capable of adjusting the injected noise rate through the use of corruption tags, as detailed in the methodology section. To this end, we insert corruption tags at various rates to simulate increasing amounts of noise. We then record the resulting transcription quality and summarization performance for each noise level. As depicted in Figure \ref{fig:noise_vs_quality}, we observe a clear trend: as the noise rate increases, the quality of the generated summaries decreases. This is evidenced by a gradual drop in the summarization quality quantified by the metric scores. This demonstrates the controllability of our noise generation method, which is essential for adapting the synthetic noise generation to errors made by various different ASR models.
\section{Conclusion}
This study highlights the critical role of Automated Speech Recognition (ASR) technology in transcribing medical dialogues and the consequent impact of ASR errors on downstream summarization tasks. Recognizing the challenges posed by limited availability of supervised data, we explored a novel approach to improve summarization models through data augmentation. By leveraging large language models (LLMs) to generate synthetic noisy dialogues that accurately reflect real-world ASR errors, we developed a method for enhancing robustness in clinical dialogue summarization. Our findings demonstrate that LLMs can effectively model ASR noise, and the use of these noisy transcripts for data augmentation leads to significant improvements in the performance of summarization models.

\section*{Acknowledgements}
Viktor is supported by the National Research Foundation, Prime Minister’s Office, Singapore under its Campus for Research Excellence and Technological Enterprise (CREATE) programme.

\bibliography{aaai25}

\end{document}